\documentclass[letterpaper, 10 pt, conference]{ieeeconf}  

\IEEEoverridecommandlockouts                              

\usepackage{amsmath} 
\usepackage{amssymb}  
\usepackage{graphicx} 
\usepackage{multirow}
\usepackage{booktabs}
\usepackage{balance}
\usepackage{xcolor,colortbl}
\usepackage{hyperref}
\usepackage{xfrac}
\usepackage{cite}

\definecolor{lavender}{rgb}{0.9, 0.9, 0.98}
\definecolor{red}{RGB}{255, 0, 0}
\definecolor{orange}{RGB}{252, 130, 62}
\definecolor{blue}{RGB}{0, 0,255}
\definecolor{darkgreen}{RGB}{0, 150,0}
\newcommand{\note}[1]{\color{orange}}

\title{\LARGE \bf
Multi-Agent Obstacle Avoidance using Velocity Obstacles \\ and Control Barrier Functions
}

\author{Alejandro Sánchez Roncero, Rafael I. Cabral Muchacho and Petter \"Ogren
\thanks{This work was partially supported by the Wallenberg AI, Autonomous Systems and Software Program (WASP) funded by the Knut and Alice Wallenberg Foundation. The authors are with the Robotics, Perception and Learning Lab., School of Electrical Engineering and Computer Science, Royal Institute of Technology (KTH), SE-100 44 Stockholm, Sweden
{\tt\small alesr@kth.se, ricm@kth.se, petter@kth.se} 
Digital Object Identifier (DOI): see top of this page.
Code: \href{https://github.com/KTH-RPL/MA-CBF-VO}{\texttt{https://github.com/KTH-RPL/MA-CBF-VO}}
Video: \href{https://youtu.be/Ox8v2s17gLw}{\texttt{https://youtu.be/Ox8v2s17gLw}}        
} 
}

\begin{document}

\maketitle
\thispagestyle{empty}
\pagestyle{empty}

\begin{abstract}

Velocity Obstacles (VO) methods form a paradigm for collision avoidance strategies among moving obstacles and agents.
While VO methods perform well in simple multi-agent environments, 
they do not guarantee safety and can show overly conservative behavior in common situations.
In this paper, we propose to combine a VO strategy for guidance with a Control Barrier Function approach for safety, which 
overcomes the overly conservative behavior of VOs and formally guarantees safety. We validate our method in a baseline comparison study, using second-order integrator and car-like dynamics. Results support that our method outperforms the baselines with respect to path smoothness, collision avoidance, and success rates.

\end{abstract}

\section{Introduction}


In this paper we propose a combination of Velocity Obstacles (VO) and Control Barrier Functions (CBF) for multi-agent collision avoidance. 
We use the classical CBF formulation of continuously solving an optimization problem to get a control input that is close to a desired one, while still guaranteeing safety. 
Instead of treating the VO as a constraint in this optimization, as~\cite{tayalControlBarrierFunctions2024}, we consider it in the objective function, as suggested in~\cite{kimStudyOptimalVelocity2016}, while keeping a less restrictive CBF as a constraint to guarantee safety.

Multi-agent collision avoidance is a well-studied problem with important applications in automated warehouses, autonomous driving, and airborne drone delivery systems.
The idea of VO was first proposed in~\cite{fioriniMotionPlanningDynamic1998}, with a number of refinements in e.g. \cite{vandenbergReciprocalVelocityObstacles2008,kimStudyOptimalVelocity2016,tayalControlBarrierFunctions2024}, and a nice recent survey in~\cite{vesentiniSurveyVelocityObstacle2024}.
The key idea is to make the assumption that all other agents, at least temporarily, will keep a constant velocity, make sure your own velocity is such that no collisions occur, and keep repeating this as the world changes. 
This family of approaches have been shown to be very efficient in handling challenging scenarios with many agents involved.
However, as was noted in \cite{kimStudyOptimalVelocity2016}, the approach might be overly conservative in some scenarios, where it leads to the conclusion that there are no admissible velocity choices left. 

A standard way of resolving this is to introduce a time horizon, where only potential collisions within that horizon are considered. This might lead to sudden changes in the control, and it was argued in 
\cite{kimStudyOptimalVelocity2016} that it makes more sense to move the VO into the objective of an optimization, where actions leading to collisions are penalized,  with higher weight given to more urgent cases.
However, none of the classical VO papers include safety guarantees with respect to collisions.
This problem is underlined by the 
numerical investigations of \cite{douthwaiteVelocityObstacleApproaches2019}, where none of the VO approaches manages to avoid collisions in all scenarios.

\begin{figure}[t]
    \centering
    \includegraphics[width=\linewidth]{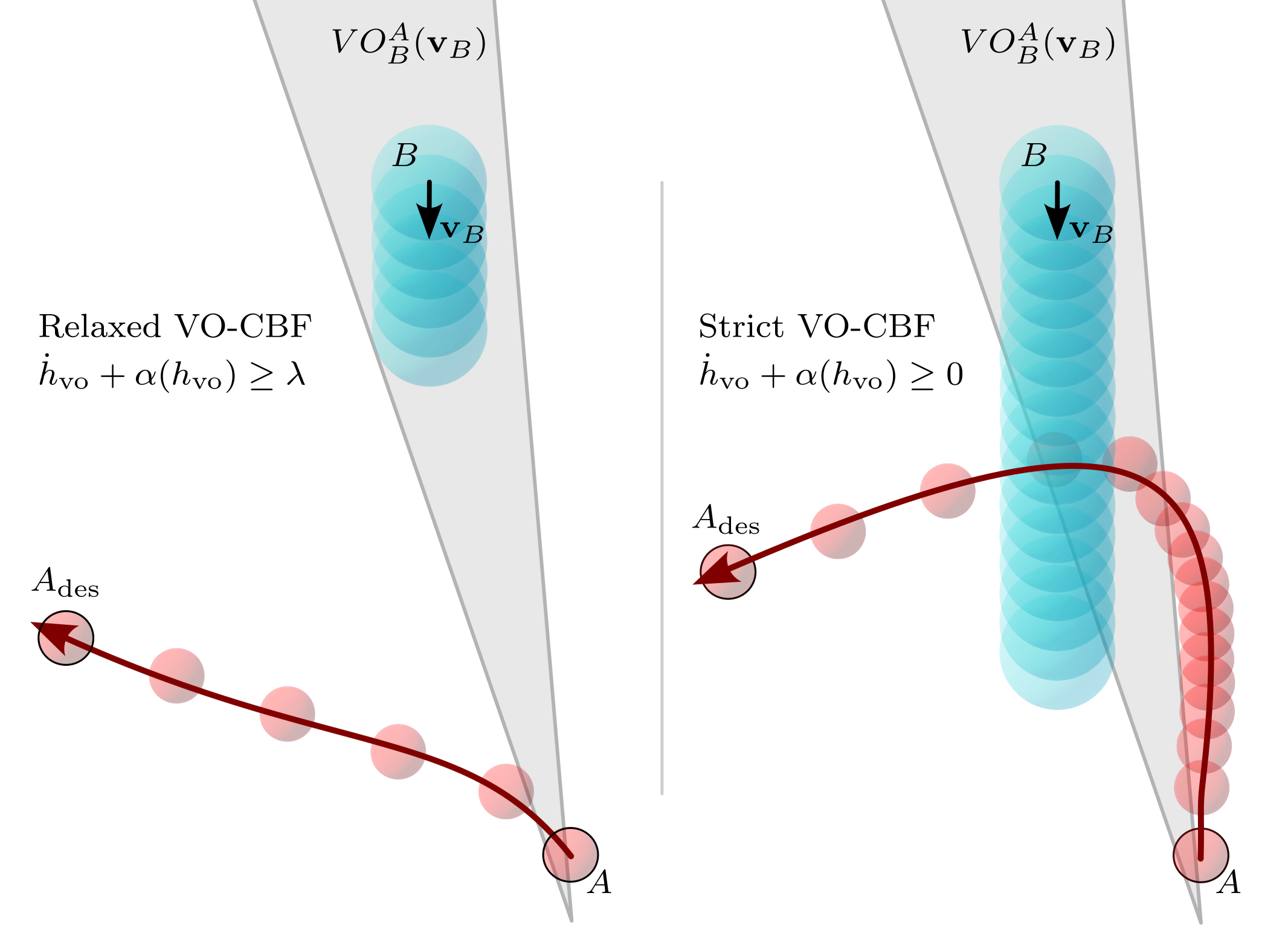}
    \caption{
    An illustration showing that strict VO-based constraints can be overly conservative. Two red agents $A$ aim for a destination $A_\mathrm{des}$ while avoiding the obstacle $B$: the one with a relaxed constraint (left) follows a direct path, while the one with a strict constraint (right) cannot turn in front of the obstacle (whose initial VO is shown in light gray) as this would imply at some point selecting a velocity inside the VO. In more crowded scenarios, such strict constraints can severely limit available options.
    }  
    \label{fig:vocbf_relax_strict}
\end{figure}

An efficient tool for guaranteeing safety in autonomous robotics is CBFs \cite{amesControlBarrierFunctions2019}.
The central idea in CBFs is to formulate the safety constraint in terms of a scalar function, and then control the time derivative of that function, in a way similar to control Lyapunov functions \cite{sastry2013nonlinear} so that the system never reaches the un-safe region.

Combining VO and CBFs is a natural idea, and has been investigated in \cite{tayalControlBarrierFunctions2024,tayalPolygonalConeControl2024,thontepuCollisionConeControl2023}.
In these papers, the VO was included in the constraints of the CBF. However, as noted by \cite{kimStudyOptimalVelocity2016}, see above, this might exclude a large part of the available control options in a scenario with many agents.

The main contribution of this paper is that we combine the ideas of \cite{tayalControlBarrierFunctions2024} and \cite{kimStudyOptimalVelocity2016}
by merging CBFs and VO, but not by adding the VO as a hard constraint, but by using a slack variable to bring the VO into the objective function, while keeping another, less restrictive
CBF constraint to guarantee safety.

The outline of the paper is as follows.
In Section~\ref{sec_background} we provide a brief background on CBFs and VO, and related work within the area can be found in Section~\ref{sec_related_work}.
Then, in Section~\ref{sec_proposed_approach} we present the proposed approach and the theoretical safety guarantees.
The simulation results can be found in Section~\ref{sec_simulations}, and conclusions are summarized in Section~\ref{sec_conclusions}.









\section{Background}
\label{sec_background}
In this section, we describe the two approaches that we seek to combine. 
First, we review the core result of CBFs, and then the VO approach. 
Finally, we briefly describe the motion models used in the examples.

\subsection{Control Barrier Functions}
CBFs have their roots in Lyapunov Theory \cite{sastry2013nonlinear}, and an overview can be found in \cite{amesControlBarrierFunctions2019}.
Let the system dynamics be control affine, i.e.,
\begin{equation}
    \dot{\mathbf{x}} = \mathbf{f}(\mathbf{x}) + \mathbf{g}(\mathbf{x})\mathbf{u},
    \label{eq:x_dot}
\end{equation}
where $\mathbf{x} \in \mathbb{R}^n$, $\mathbf{u} \in \mathbb{R}^m$, $\mathbf{f}:\mathbb{R}^n \rightarrow \mathbb{R}^n$ and $\mathbf{g}:\mathbb{R}^n \rightarrow \mathbb{R}^{n \times m}$.

Let the safe set be given by
$\mathcal{C}=\{\mathbf{x}: h(\mathbf{x}) \geq 0\}$,
where $h$ depends only on the state 
and has to satisfy some properties.
Then, given the system dynamics $\mathbf{f}(\mathbf{x}, \mathbf{u})$, if the set 
\begin{align}
\label{eq:CBF}
K = \{\mathbf{u} \in U: \frac{dh}{d\mathbf{x}}(\mathbf{f}(\mathbf{x})+\mathbf{g}(\mathbf{x})\mathbf{u}) \geq - \alpha(h(\mathbf{x}))\},
\end{align}
is non-empty for all $\mathbf{x}$, and if we choose controls $\mathbf{u}$ inside $K$,
we are guaranteed to stay in the safe set $\mathbf{x} \in \mathcal{C}$, see~\cite{amesControlBarrierFunctions2019}. The function $\alpha$ is a so-called class $ \mathcal{K}$ function, that is 
${\alpha: \mathbb{R}_+ \rightarrow  \mathbb{R}_+,}\, {\alpha(0)=0,}$ and $\alpha$ is strictly monotonic increasing, see Theorem 2 in \cite{amesControlBarrierFunctions2019}. Also note that sometimes the Lie-derivative notation is used, where
\begin{align}
\frac{dh}{d\mathbf{x}}\dot{\mathbf{x}} = \frac{dh}{d\mathbf{x}}(\mathbf{f}(\mathbf{x})+\mathbf{g}(\mathbf{x})\mathbf{u}) = L_{\mathbf{f}} h(\mathbf{x}) + L_{\mathbf{g}} h(\mathbf{x}) \mathbf{u}.
\end{align}

If some desired control is given by $\mathbf{u}_\mathrm{des}=\mathbf{k}(\mathbf{x})$,
we can formulate the following optimization problem to find a control $\mathbf{u}$ that is close to $\mathbf{k}(\mathbf{x})$ while still inside $K$, thereby keeping the state inside the safe set $\mathcal{C}$,
\cite{amesControlBarrierFunctions2019}
\begin{align}
    \mathbf{u}(\mathbf{x}) = \operatorname*{argmin}_{\mathbf{u}} &\quad  \frac{1}{2}|| \mathbf{u} - \mathbf{k}(\mathbf{x})   ||^2 \\
    \mbox{s.t.} &\quad  \mathbf{u} \in K.
\end{align}
The key observation about the problem above is that it is a so-called Quadratic Programming (QP) problem (Linear constraints and quadratic objective function) which can be solved to optimality very efficiently,  allowing the QP-solution to be part of a closed control loop.

\subsection{Velocity Obstacles}
\begin{figure}
    \centering
    \includegraphics[width=8cm]{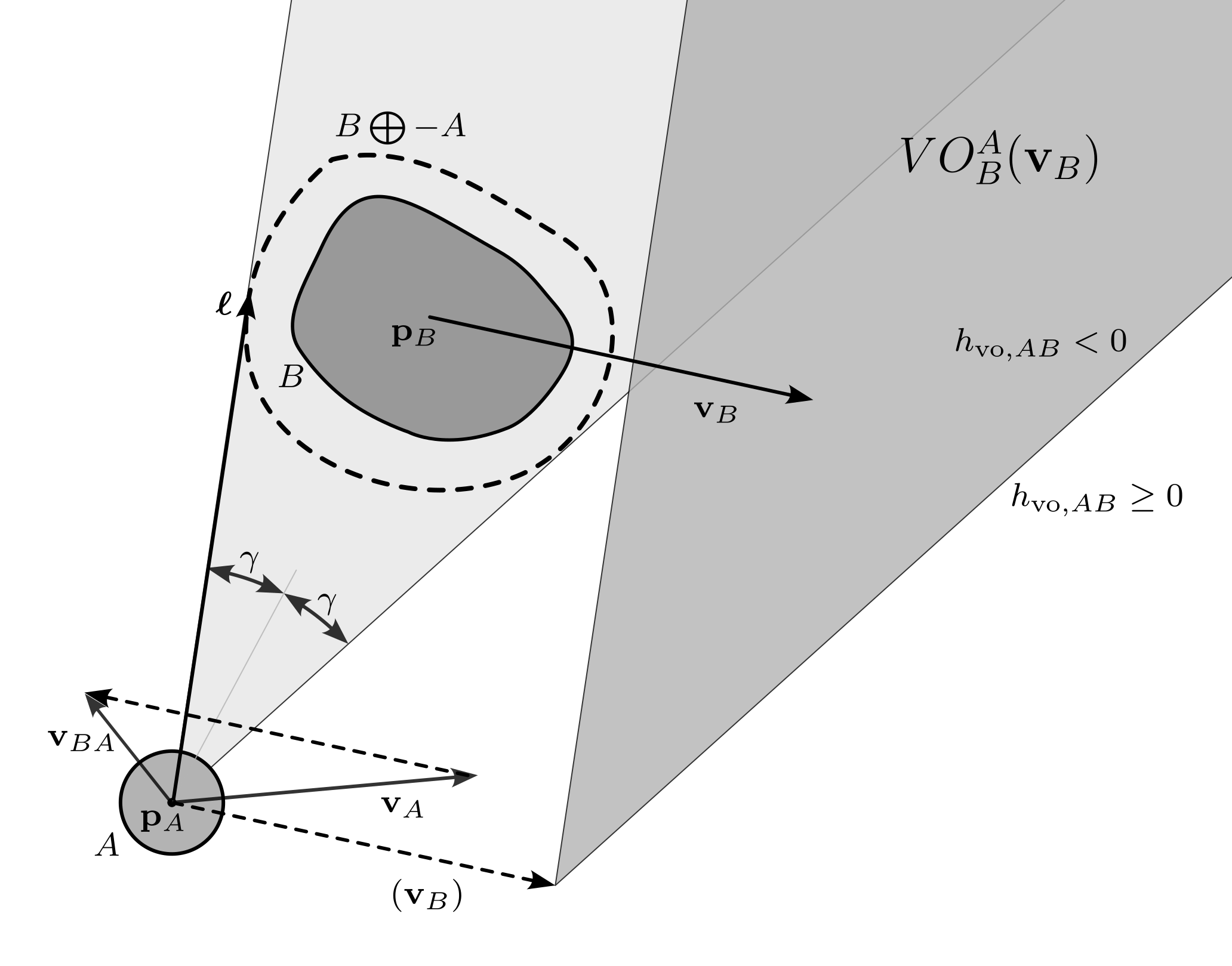}
    \caption{Illustration of the key idea and notation used in Velocity Obstacle methods. If agent $A$ chooses a velocity $\mathbf{v}_A$ outside the dark cone $VO_B^A(\mathbf{v}_B)$, it will not collide with agent $B$.}
    \label{fig_VO}
\end{figure}

The idea of Velocity Obstacles (VO) 
was introduced in \cite{fioriniMotionPlanningDynamic1998} and refined in e.g. 
\cite{vandenbergReciprocalVelocityObstacles2008,kimStudyOptimalVelocity2016,tayalControlBarrierFunctions2024}. 
The key idea is illustrated in Figure \ref{fig_VO}.
If agent $B$ is moving with fixed velocity, we can study the avoidance problem from the perspective of agent $A$.
If agent $A$ chooses the same velocity as agent $B$ (at the lower tip of the darker cone) the relative distance will not change. If a component is then added moving towards $B$, there will be a collision at some point. The higher up in the dark cone, the sooner the collision will happen (we might think of the lower tip as a collision at $t=\infty$). 

Formally, allowing arbitrary shapes for $A$ and $B$ by using the $\oplus$-notation, we can describe the concept above as follows.
First, we define some set operations for $A,B \subset \mathbb{R}^n$ and $\mathbf{p},\mathbf{v} \in \mathbb{R}^n$
\begin{align}
   A \oplus B &= \{\mathbf{a} + \mathbf{b} \mid \mathbf{a} \in A, \mathbf{b} \in B\} \\
   -A &= \{-\mathbf{a} \mid \mathbf{a} \in A\} \\
   \tau(\mathbf{p},\mathbf{v})&= \{\mathbf{p} + t\mathbf{v} \mid t \geq 0\},
\end{align}
where $\tau(\cdot)$ maps a state $(\mathbf{p}, \mathbf{v})$ to its future trajectory assuming constant velocity. 
Let two agents have positions $\mathbf{p}_A, \mathbf{p}_B$ and velocities $\mathbf{v}_A, \mathbf{v}_B$, all in $\mathbb{R}^n$.
The velocity obstacle caused by $B$ from the perspective of $A$ is thus
${VO_B^A(\mathbf{v}_B) = \{\mathbf{v}_A \mid \tau(\mathbf{p}_A, \mathbf{v}_A - \mathbf{v}_B) \cap B \oplus -A \neq \emptyset  \}
}$.

\subsection{Vehicle motion models used in the simulations}

Here we describe the different model dynamics considered in this work. It should be noted that other models might equally work and have not been considered here.

For the double integrator in $2$ dimensions we have ${\mathbf{x}=(p_1, p_2, v_1, v_2)^T \in \mathbb{R}^{4}}$ and ${\mathbf{u} \in \mathbb{R}^{2}}$, giving the affine form
\begin{equation}
    \dot{\mathbf{x}} = \mathbf{f}(\mathbf{x}) + \mathbf{g}(\mathbf{x})\mathbf{u},
\end{equation}
with 
\begin{equation}
    \mathbf{f}(\mathbf{x}) = \begin{bmatrix} 
   v_1 \\
   v_2 \\
   0 \\
   0
    \end{bmatrix}, \quad
     \mathbf{g}(\mathbf{x}) = 
 \begin{bmatrix} 
   0 & 0 \\
   0 & 0 \\
   1 & 0 \\
   0 & 1 
    \end{bmatrix}.
\label{eq:second_integrator}
\end{equation}

For the car we have
\begin{align}
    \dot p_1 &= v \cos \theta \\
    \dot p_2 &= v \sin \theta \\
    \dot \theta &= v/L \tan \phi \\
    \dot v &= a,
\end{align}
where $p_1, p_2$ is the position, $\theta$ represents the orientation of the car, $v$ is the speed, $L$ is the wheelbase distance, $\tan \phi$ is the steering control, and $a$ is the acceleration control.
Writing the above in the affine form ${\dot{\mathbf{x}} = \mathbf{f}(\mathbf{x}) + \mathbf{g}(\mathbf{x})\mathbf{u}}$,
with ${\mathbf{x}=(p_1, p_2, \theta, v)}$, and ${\mathbf{u}=(\tan \phi, a)}$ we get
\begin{equation}
    \mathbf{f}(\mathbf{x}) = \begin{bmatrix} 
   v \cos \theta \\
   v \sin \theta \\
   0 \\
   0
    \end{bmatrix}, \quad
     \mathbf{g}(\mathbf{x}) = 
 \begin{bmatrix} 
   0 & 0 \\
   0 & 0 \\
   v/L & 0 \\
   0 & 1 
    \end{bmatrix}.
    \label{eq:car_dynamics}
\end{equation}
Note that we treat $\tan \phi$ as the control variable to avoid having a nonlinear expression in $\mathbf{u}$.








\section{Related Work}
\label{sec_related_work}

Collision avoidance in multi-agent systems is a vast field in itself, with subfields such as path planning \cite{desaraju2012decentralized}, model predictive control (MPC) \cite{frasch2013auto,piccinelliMPCBasedMotion2023} and artificial potential functions \cite{khatib1986potential}.
This paper is about merging the ideas of VO and CBFs so we will focus on the description of related work in these two areas.

The idea of velocity obstacles was introduced in 
\cite{fioriniMotionPlanningDynamic1998} and extended in a number of papers \cite{van2011reciprocal,snape2011hybrid,vandenbergReciprocalVelocityObstacles2008,alonso2013optimal}, that have refined and developed the concept since.
One identified issue with VO was the oscillation problem.
 If two agents are facing each other, then they might first select a velocity that avoids collision. But in the next timestep, they might conclude that no collision is imminent, given the current velocity of the other agent, and then (both) turn back to their initial velocities.
  This will create an oscillation and works such as the Reciprocal Velocity Obstacle (RVO)
   \cite{van2011reciprocal} were proposed to resolve it.
   The RVO was later extended into the Hybrid RVO by \cite{snape2011hybrid} reducing the problem further, by making the velocity obstacles asymmetric, and built upon in 
   \cite{vandenbergReciprocalVelocityObstacles2008,alonso2013optimal} where the problem is reduced to a linear programming problem that can be efficiently solved.
   In a paper on combining VO with MPC \cite{piccinelliMPCBasedMotion2023}, the idea of including VO as constraints in the short-horizon motion planning problem was explored.

The work that lies closest to our approach within the VO literature is \cite{kimStudyOptimalVelocity2016}, where a VO term is used as part of the objective.
We believe that this is a very good design choice, but our approach goes beyond 
\cite{kimStudyOptimalVelocity2016} in the sense that we include a CBF element (not relying on VO) that guarantees safety.

CBFs are a tool for guaranteeing safety in control systems.
Early works include \cite{amesControlBarrierFunction2017,ogrenAutonomousUCAVStrike2006}, with an overview in \cite{amesControlBarrierFunctions2019}, and an extension to higher order systems in \cite{xiaoControlBarrierFunctions2019}.
In connection to the avoidance of static obstacles, the connection between CBFs and artificial potential functions was explored in \cite{singletaryComparativeAnalysisControl2021}.



The works closest to ours within the CBF literature are~\cite{thontepuCollisionConeControl2023} and \cite{tayalControlBarrierFunctions2024}, where the idea of combining CBFs and VO is proposed. 
However, in both papers, the authors add the VO as a hard constraint, which can be overly restrictive, as observed in \cite{kimStudyOptimalVelocity2016} and illustrated in Fig. \ref{fig:vocbf_relax_strict}. 
Thus our approach goes beyond \cite{thontepuCollisionConeControl2023,tayalControlBarrierFunctions2024} in the sense that the VO constraint is relaxed, and thereby essentially moved into the objective of the optimization, while another (non-VO) collision avoidance component is added to the CBF to provide the safety guarantees.

\section{Method}
\label{sec_proposed_approach}
    Instead of using velocity obstacles to directly ensure safety, we employ them for guidance through a relaxed inequality constraint and use a valid CBF to guarantee safe behavior, under the assumption that all agents seek to avoid collisions. Here, agent $i$ denotes the agent for which we want to solve the optimization problem, and $j$ indexes the dynamic obstacles. Altogether, the optimization problem for the agent $i$ is formulated as
\begin{align}
    \mathbf{u}_i = \operatorname*{argmin}_{\mathbf{u}_i} &\quad k_u \lVert \mathbf{u}_i - \mathbf{u}_{\mathrm{ref}, i} \rVert^2 + k_\mathrm{vo} \sum_{j=1}^{n_\mathrm{obs}} w_{ij} \lambda_{ij}^2
    \label{eq_main_objective}
    \\
    \mbox{s.t.} &\quad \dot{h}_{\mathrm{vo}, ij} + \alpha_\mathrm{vo}(h_{\mathrm{vo}, ij}) \geq \lambda_{ij}, \label{eq:main_h_vo}\\
    &\quad \dot{h}_{c, ij} + \alpha_c (h_{c, ij}) \geq 0 \label{eq:main_h_c},\\
    &\quad  \mathbf{u}_i \in \mathcal{U}, \label{eq:main_u}
\end{align}
where $\mathcal{U} = \{ \mathbf{u} \in \mathbb{R}^m \mid \|\mathbf{u}\| \leq u_{\mathrm{max}} \}$ represents the valid input set. In this section we describe in detail the functions and variables composing the above optimization problem, starting with the objective function~\eqref{eq_main_objective} and then finishing with the constraints~\eqref{eq:main_h_vo}$-$\eqref{eq:main_u} guaranteeing safety.

\subsection{Objective Function}

The objective function for each agent is composed of a goal attractive component $J_\mathrm{goal}$ and a velocity-obstacle guidance component $J_\mathrm{vo}$.
The goal attractive component is defined as the squared norm of the difference between the optimization variable to a reference input  
\begin{align}
    J_{\mathrm{goal}, i} = \lVert \mathbf{u}_i - \mathbf{u}_{\mathrm{ref}, i} \rVert^2,
\end{align}
where the reference input $\mathbf{u}_\mathrm{ref}$ is given or computed by a known and state-dependent policy.
In a scenario that is challenging with respect to the static obstacles, this would include planning a path that leads to the goal and then tracking this path. 
The path tracking control would then be $\mathbf{u}_\mathrm{ref}$.

The velocity obstacle component is inspired by the VO-CBF constraint~\cite{tayalPolygonalConeControl2024} 
\begin{align}    
    h_{\mathrm{vo}, ij} (\mathbf{x}) &=  \mathbf{p}_{ij}^T  \mathbf{v}_{ij} + \|\mathbf{p}_{ij}\| \|\mathbf{v}_{ij}\| \cos(\gamma_{ij}) \label{eq:h_vo} \\ 
    \dot{h}_{\mathrm{vo}, ij} &\geq - \alpha_{\mathrm{vo}}(h_{\mathrm{vo}, ij}), \label{eq:cbf_vo_constraint}
\end{align}
where $\mathbf{p}_{ij} = \mathbf{p}_j - \mathbf{p}_i $ and $ \mathbf{v}_{ij} = \mathbf{v}_j - \mathbf{v}_i$ represent the relative position and velocity vectors from agent $i$ to $j$ respectively,
the angle $\gamma$ is the semi-angle of the velocity obstacle cone with $\cos(\gamma_{ij}) = \hat{\mathbf{p}}_{ij}^T\hat{\boldsymbol{\ell}}_{ij}$, where \textit{hat} denotes the normalized vector. 
The vector $\boldsymbol{\ell}_{ij}$ points from the $i$ object's center, to a point on the augmented $j$ object's surface and on the corresponding cone.
The notation is also visualized in Figure~\ref{fig_VO}. 
The time derivative of $h_{\mathrm{vo}, ij}$, in  \eqref{eq:h_vo}, removing object subscripts, is given by
\begin{align}
    \dot{h}_{\mathrm{vo}} &= 
    \dot{\mathbf{p}}^T\mathbf{v} + 
    \mathbf{p}^T\dot{\mathbf{v}} +
    \dot{\mathbf{v}}^T\hat{\mathbf{v}} \mathbf{p}^T \hat{\boldsymbol{\ell}} + 
    \|\mathbf{v}\| \dot{\mathbf{p}}^T \hat{\boldsymbol{\ell}}
    + \|\mathbf{v}\| \mathbf{p}^T \dot{\hat{\boldsymbol{\ell}}} \notag \\ 
     &=  
    \mathbf{u}^T (\mathbf{p} + \hat{\mathbf{v}} \mathbf{p}^T \hat{\boldsymbol{\ell}} ) + 
    \|\mathbf{v}\| (\mathbf{v}^T \hat{\boldsymbol{\ell}}
    +  \mathbf{p}^T \dot{\hat{\boldsymbol{\ell}}} + \|\mathbf{v}\|),
\end{align}
which is linear in acceleration $\mathbf{u}$ for the double integrator system.

Intuitively, the cone CBF constraint \eqref{eq:cbf_vo_constraint} keeps the relative vector $\mathbf{v}_{ji}$ outside of the Velocity Obstacle cone.
Thus, satisfying (\ref{eq:cbf_vo_constraint}) keeps the system safe, but it is overly constraining in many real scenarios, as pointed out in \cite{kimStudyOptimalVelocity2016}, and illustrated in  Figure~\ref{fig:vocbf_relax_strict}.
As can be seen, this problem is significant in the case of a single obstacle and gets even worse with multiple obstacles around, potentially making it very 
difficult to find a velocity outside of the joint set of Velocity Obstacles. 
The basic formulation used in \cite{thontepuCollisionConeControl2023} would result in the agent not finding any solution to the set of inequalities (other than the trivial 0 if its velocity is 0). 
To overcome this problem, we incorporate the VO as part of the objective by introducing an  auxiliary (slack) variable $\lambda$
in Equation (\ref{eq:cbf_vo_constraint}), giving
\begin{align}  \dot{h}_{\mathrm{vo}, ij} + \alpha_\mathrm{vo}(h_{\mathrm{vo}, ij}) \geq \lambda_{ij}.
\end{align}
and then penalizing this slack variable in the objective as
\begin{align}    J_{\mathrm{vo},ij} = \lambda_{ij}^2. 
\end{align}

The total objective function now becomes a weighted sum of the objective components, 
\begin{align}
    J_i = k_u J_{\mathrm{goal}, i} + k_\mathrm{vo} \sum_{j=1}^{n_\mathrm{obs}} w_{ij} J_{\mathrm{vo}, ij}, \label{eq:total_objective}
\end{align}
with constant positive scaling factors $k_u, k_\mathrm{vo} \in \mathbb{R}_+$. The weights are computed as a function of an approximate time-to-collision $T_{col, ij}$ (based on the linear extrapolation of positions) between objects $i$ and $j$
\begin{align}
    w_{ij} = \frac{1}{T_{\mathrm{col},ij}}.
\end{align}
If objects $i$ and $j$ do not collide according to the used model, then $T_{\mathrm{col},ij}\rightarrow \infty$ and we set $w_{ij} = 0$. When using a linear model for extrapolation, the time-to-collision and therefore the weights are symmetric, i.e., $w_{ij}=w_{ji}$. The choice of an inverse time-to-collision for weighting encodes that an agent should dedicate more effort to avoid objects with which it would collide sooner if no further action is taken. 

\subsection{Safety-Critical Constraint}

The safety of agents is formally guaranteed by a CBF based on position and velocity for collision avoidance~\cite{ghaffari2018safety, chen2021backup}. 
The function $h_c$ implicitly describes safe states as states at which the distance to an obstacle is higher than the braking distance at maximum acceleration or braking effort.

The variable $d_{ij}$ represents the minimum distance between the $i$ and $j$ objects. 
We assume objects are convex, and therefore each of their distance functions is continuously differentiable.
This assumption is not restrictive as continuously differentiable approximations of distance functions are readily available~\cite{muchacho2024adaptive, li2024representing, liu2022regularized}.
A valid CBF for double integrator systems with saturated input $\lVert \mathbf{u} \lVert \leq u_\mathrm{max}$ is
\begin{align}
    h_{c,ij} = d_{ij} - \delta - \frac{\nu_{ij}^2}{2u_\mathrm{max}} 
    \label{eq:h_brake} \\
    \nu_{ij} = \min\left(0, \mathbf{v}_{ij}^T\hat{\mathbf{p}}_{ij}\right),
\end{align}
where 
$\delta \geq 0$ is a safety margin,
$\hat{\mathbf{p}}_{ij}$ is the normalized vector between the agent and the closest point on object $j$, which makes $\nu_{ij}$ the relative velocity of the other object projected onto this direction with large negative $\nu_{ij}$ indicating a potential problem. 

 In the case where $\mathbf{v}_{ij}^T\hat{\mathbf{p}}_{ij} \leq 0$
 we get
\begin{align}
    \dot{h}_{c,ij} &= \dot{d}_{ij} - \frac{\nu_{ij} 
    }{u_\mathrm{max}} ( \mathbf{u}_{ij}^T\hat{\mathbf{p}}_{ij} + \mathbf{v}_{ij}^T\dot{\hat{\mathbf{p}}}_{ij})
\end{align}
i.e., time derivative of the function \eqref{eq:h_brake} is linear in the acceleration.
If $\mathbf{v}_{ij}^T\hat{\mathbf{p}}_{ij} > 0$ we get $\dot{h}_{c,ij} = \dot{d}_{ij}=\mathbf{v}_{ij}^T\hat{\mathbf{p}}_{ij}>0$.
Either way, the corresponding linear inequality constraint to guarantee safety will be
\begin{align}
    \dot{h}_{c, ij}(\mathbf{u}_{ij}) + \alpha_c (h_{c, ij}) \geq 0, \label{eq:hcineq}
\end{align}
 where  $\alpha_c$ is a class $ \mathcal{K}$ function. Note that if $\mathbf{v}_{ij}^T\hat{\mathbf{p}}_{ij} > 0$, Equation \eqref{eq:hcineq} is independent of $\mathbf{u}_{ij}$ and always satisfied.

\section{Experiments}
\label{sec_simulations}

In this section, we evaluate our method in multi-agent scenarios, through a baseline comparison study and in a validation scenario with car-like dynamics. We show that 
\begin{enumerate}
    \item the formal safety guarantee translates into safe agent behavior in practice,
    \item the VO-based component successfully guides the agents through the tasks,
    \item our method outperforms baselines w.r.t. path smoothness, collision avoidance, and success rates.
\end{enumerate}
Code\footnote{\href{https://github.com/KTH-RPL/MA-CBF-VO}{\texttt{https://github.com/KTH-RPL/MA-CBF-VO}}} and videos\footnote{\href{https://youtu.be/Ox8v2s17gLw}{\texttt{https://youtu.be/Ox8v2s17gLw}}} of the  experiments are available online.


\begin{figure*}[]
\begin{center}
  \includegraphics[width=0.9\textwidth]{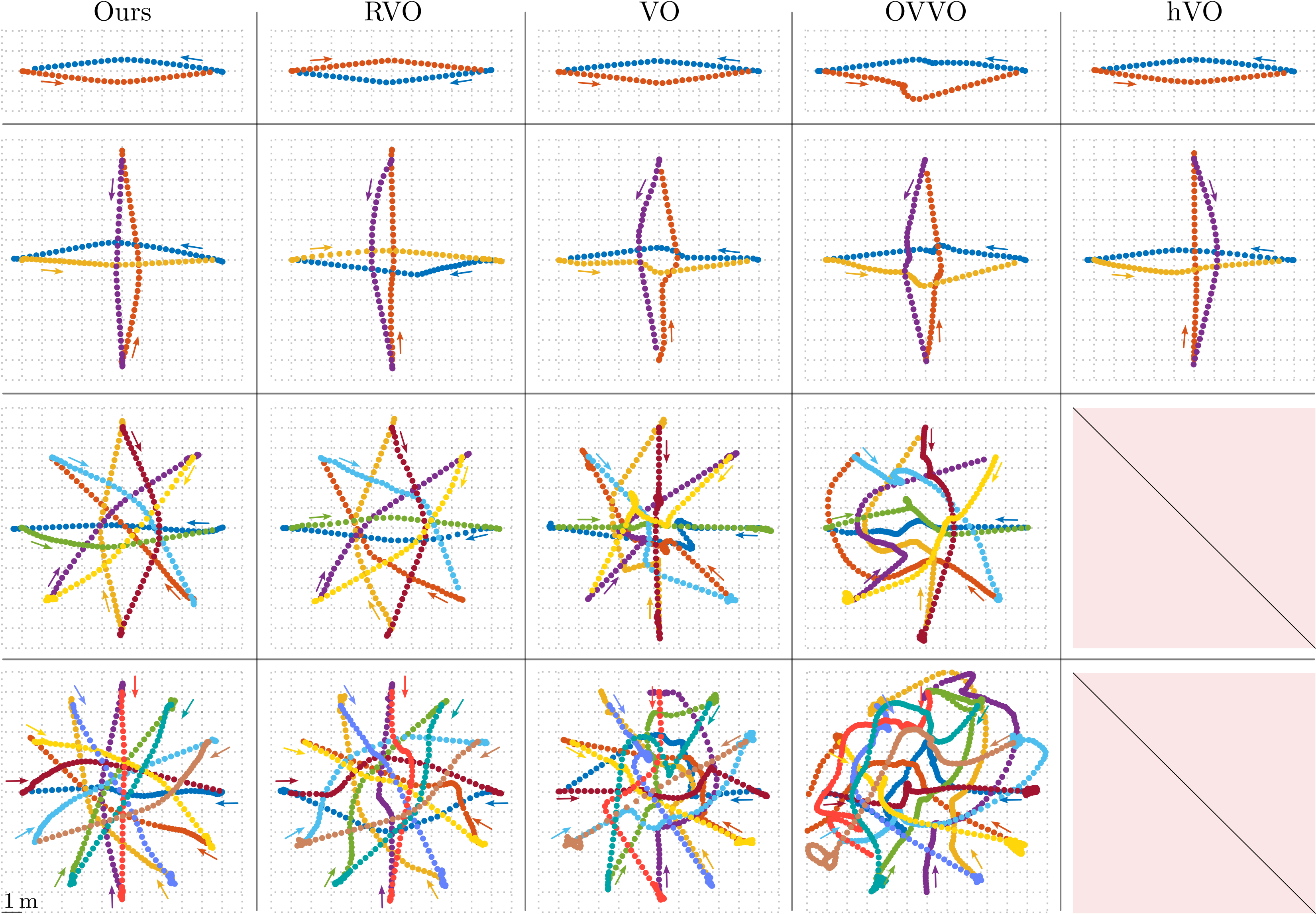}
\end{center}
  \caption{Visualization of the resulting trajectories in the baseline comparison study. 
  The arrows in each corresponding color describe the direction of motion of each agent.
  Our method (left-most column) remains stable and computes smooth trajectories for all evaluated scenarios.
  The hVO method encountered infeasibility (no available safe velocities) in the scenarios with $8$ and $12$ agents. 
  The aspect ratio is equal for both dimensions and the scale is given by the bottom left reference.
  }
  \label{fig:baseline_result_figure}
\end{figure*}

\subsection{Setup} \label{sec:exp_setup}

Our approach is implemented in MATLAB and Simulink3D. 
For simplicity, agents are modeled as spheres with constant radii. 
We evaluate both 2nd-order integrator dynamics and car-like dynamics, given by Equations~\eqref{eq:second_integrator}, \eqref{eq:car_dynamics} respectively.
We assume full knowledge of the environment, including the positions, velocities, and dimensions of the agents. 
While the approach can be extended to handle more general agent shapes or uncertainties in position and velocity, see \cite{snape2011hybrid}, we focus here on deterministic settings. 

\begin{table}[]
\centering
\begin{tabular}{@{}ccc@{}}
\toprule
\multirow{2}{*}{Parameter}       & \multirow{2}{*}{Second-order Int.} & \multirow{2}{*}{Car Dynamics} \\
                                &         &         \\ \midrule
Simulation time                 & $60\,\mathrm{s}$  & $60\,\mathrm{s}$  \\
Timestep                        & $10\,\mathrm{ms}$ & $10\,\mathrm{ms}$ \\
$\alpha_{\mathrm{vo}}$                          & $10$      & $10$      \\
$\alpha_{\mathrm{c}}$                        & $10$      & $10$      \\
$k_u$                           & $1$       & $1$       \\
$k_\mathrm{vo}$                 & $1000$    & $1$       \\
Preferred velocity ($v_\mathrm{pref}$) & $1\,\mathrm{m/s}$ &  -       \\
Maximum velocity ($v_\mathrm{max}$)     & $2\,\mathrm{m/s}$ & $10\,\mathrm{m/s}$ \\
Maximum acceleration ($u_\mathrm{max}$) & $1\,\mathrm{m/s^2}$ & $3\,\mathrm{m/s^2}$ \\
Maximum steering ($\tan(\phi)$) & -       & $2$       \\
Geometric tolerance             & $10\,\% $   & $10\,\%$    \\
Goal position tolerance         & $0.5\,\mathrm{m}$ & $1\,\mathrm{m}$ \\
Agent Radii                     & $0.5\,\mathrm{m}$ & $1\,\mathrm{m}$ \\
Car's characteristic length ($L$) & $1\,\mathrm{m}$ & $1\,\mathrm{m}$ \\
P coefficient                   & $1$       & $0.2$     \\
D coefficient                   & $0.5$     & -       \\
N sampling points               & $250$     & -       \\ 
$k_{tp}$, $k_{vd}$, $c_1$, $c_2$ &   $2$, $1$, $1$, $1$ & - \\ \bottomrule
\end{tabular}
\caption{Main parameters of the simulation.}
\label{tab:simulation_settings}
\end{table}

We run simulations with fixed timesteps, with the control output calculated for each agent at each step. 
The states are updated simultaneously for all agents. 
For our model, we define constraints and solve the optimization problem using MATLAB's linear least-squares solver at each timestep. 
The acceleration is constrained during optimization, and forward Euler integration is used to propagate the resulting accelerations to compute velocities and positions. Further details on the simulation setup are provided in Table~\ref{tab:simulation_settings}.

\subsection{Comparison to Baselines} \label{sec:scenarios}

\subsubsection{Setup}

As baselines, we used Velocity Obstacles (VO), Reciprocal Velocity Obstacles (RVO), Collision Cone CBF (hVO), and Optimal Velocity selection using Velocity Obstacle (OVVO). For implementation, we followed the methodology outlined in \cite{van2011reciprocal, kimStudyOptimalVelocity2016}, while focusing on the second-order integrator dynamics. 
To handle the fact that the other approaches had no explicit motion model, with control $\mathbf{u}$, we did as follows.
At each timestep, we define and sample a set of admissible velocities based on the maximum allowable acceleration and timestep. For VO and RVO, we select the one that minimizes (analogous to (\ref{eq_main_objective}))
the penalty
\begin{equation} J_{\mathrm{VO}, \mathrm{RVO}} =  \frac{1}{T_{\mathrm{col}}} + ||\mathbf{v} - \mathbf{v}_\mathrm{des}||,
\label{eq:penalty_vo}
\end{equation}
where $T_{\mathrm{col}}$ is the minimum time-to-collision among obstacles. For OVVO, the penalty is expressed as
\begin{equation} J_{\mathrm{OVVO}} =  k_{tp} d_v^{-c_1} t_p^{-c_2} + k_{vd} ||\mathbf{v} - \mathbf{v}_\mathrm{des}||,
\label{eq:penalty_ovvo}
\end{equation}
where $\sfrac{k_{tp}}{k_{vd}}$ balances deviating from the VO cone and following the desired control input, and $c_1$, $c_2$ are constants. $d_v$ and $t_p$ represent the clearance and pass time, closely related to a separation between obstacles and time-to-collision respectively (we refer the reader to \cite{kimStudyOptimalVelocity2016} for further details). 
We recover the hVO formulation by setting $k_{cone}$ and the slack variables ($\lambda_i$) to 0 while removing the safety-critical constraints. 

The time-to-collision ($T_{\mathrm{col}}$) is computed assuming constant velocity, solving for the time $t$ such that  
\begin{align}
\tau(\mathbf{p}_A, \mathbf{v}_A^\prime - \mathbf{v}_B) \cap B \oplus -A \quad   &(\text{VO}) \\
\tau(\mathbf{p}_A, 2 \cdot \mathbf{v}_A' - \mathbf{v}_A - \mathbf{v}_B) \cap B \oplus -A \quad   &(\text{RVO}),
\end{align} 
where $\mathbf{v}_A'$ is the desired velocity for agent $A$.
The selected velocity is divided by the timestep to obtain the control input. 
All parameters of the models are manually tuned with minimal effort to evaluate qualitative performance. 
No optimization is applied. 
The control scheme is PD-based, using position error as the input to guide the agents toward their goals.

We evaluate our method in a circular formation scenario to compare its performance against the baselines. 
Agents are equidistantly positioned on a circle with zero initial velocity, and their goal is to reach the diametrically opposite point. 
This naturally creates conflicting trajectories, leading to potential collisions. 
As the system is deterministic, we add noise to the initial position to compute statistics. 
We examine cases with 2, 4, 8, and 12 agents, all governed by second-order integrator dynamics. 
Each scenario is run 10 times and statistics are reported. 

\subsubsection{Results}

Results are summarized in Table~\ref{tab:results_experiments} and visualized in Figure~\ref{fig:baseline_result_figure}.
Qualitatively, our method produces smoother trajectories compared to the baselines. While all methods perform well with few agents, increasing the number of agents resulted in greater challenges, particularly for VO, hVO, and OVVO, leading to numerous collisions or low success rates. RVO performs relatively well, but still could not completely avoid collisions, as seen in the 12-agent case. In contrast, our approach consistently yielded zero collisions, as predicted by the safety guarantees.



In terms of success rate, hVO fails at 8 and 12-agent scenarios. At the start of the simulations, all agents are in a safe state with zero velocity. 
This leads them to choose a velocity towards their goals. Thus, at the next timestep, all of them are facing each other with colliding velocities, and since the scenario is crowded and acceleration is constrained, no feasible solution is found to the problem without relaxing the constraints. 
In terms of completion time, RVO performed best, with our approach second, but never needing more than 10\% extra time to finish.

Regarding computation time, solving the quadratic programming problem in our method proved faster and more efficient than the random sampling process used by the baselines. 
However, our approach did incur some overhead due to the need to compute constraints for all agents.

Thus, while our method is similar to RVO in overall performance, it provides safety guarantees. 
Moreover, the smoother trajectories might provide less wear and tear in hardware over time.
RVO assumes a degree of cooperation among agents, whereas our method is more general and can be employed in decentralized systems. 
Since the constraints are linear, 
we get a Quadratic Programming problem (QP), keeping computational complexity unchanged—an important advantage for scaling to larger problems.

\subsection{Car Dynamics Example}

\begin{figure}[h!]
\centering
\includegraphics[width=1\linewidth]{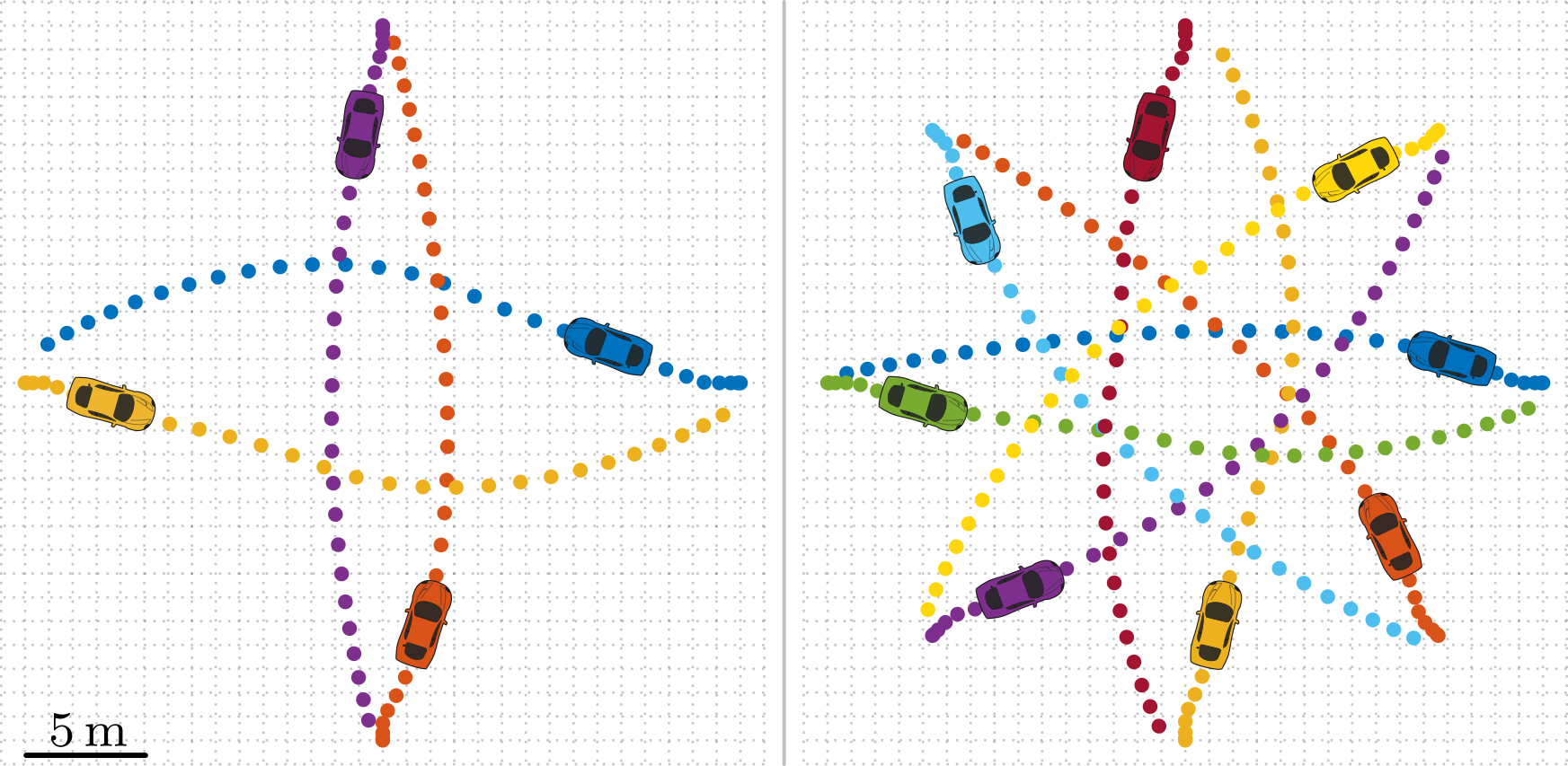}
\caption{Trajectories when applying the proposed approach to  cars.}
\label{fig:traces_cars}
\end{figure}

We extend the experiments to car-like dynamics, see Equation (\ref{eq:car_dynamics}), maintaining the same methodology. Two scenarios are evaluated: one with 4 agents and another with 8 
agents, where the agents are tasked to reach the diametrically opposite location.
The results are consistent with the second-order integrator dynamics.
As can be seen in Figure~\ref{fig:traces_cars}, the resulting trajectories are smooth,
and the non-holonomic agents successfully adapted to each other. 






\begin{table}[]
\centering
\begin{tabular}{@{}cccccc@{}}
\toprule
                     &      &                             &                                    &                     &                    \\
\multirow{-2}{*}{N.A} &
  \multirow{-2}{*}{Method} &
  \multirow{-2}{*}{S.R.} &
  \multirow{-2}{*}{Collisions} &
  \multirow{-2}{*}{\begin{tabular}[c]{@{}c@{}}Simulation\\ Time (s)\end{tabular}} &
  \multirow{-2}{*}{\begin{tabular}[c]{@{}c@{}}Computation \\ Time(ms)\end{tabular}} \\ \midrule
                     & Ours & 1                           & 0.00$\pm$0.00                          & 15.28$\pm$0.06          & 4.86$\pm$1.09          \\
                     & VO   & 1                           & 0.00$\pm$0.00                          & 15.18$\pm$0.08          & 3.84$\pm$0.70          \\
                     & RVO  & 1                           & 0.00$\pm$0.00                          & \textbf{15.13$\pm$0.08} & \textbf{3.64$\pm$0.50} \\
                     & hVO  & 1                           & 0.00$\pm$0.00                          & 15.25$\pm$0.08          & 5.00$\pm$0.00          \\
\multirow{-5}{*}{2}  & OVVO & 1                           & 0.00$\pm$0.00                          & 17.66$\pm$0.34          & 13.86$\pm$1.11         \\ \midrule
                     & Ours & 1                           & 0.00$\pm$0.00                          & 17.39$\pm$0.76          & \textbf{4.98$\pm$0.76} \\
                     & VO   & 1                           & 0.00$\pm$0.00                          & 17.31$\pm$1.36          & 10.82$\pm$0.60         \\
                     & RVO  & 1                           & 0.00$\pm$0.00                          & \textbf{15.77$\pm$0.84} & 10.32$\pm$0.72         \\
                     & hVO  & \cellcolor[HTML]{FFCCC9}0.8 & 0.00$\pm$0.00                          & 14.45$\pm$6.17          & 5.60$\pm$0.00          \\
\multirow{-5}{*}{4}  & OVVO & 1                           & 0.00$\pm$0.00                          & 30.62$\pm$8.22          & 28.24$\pm$2.30         \\ \midrule
                     & Ours & 1                           & 0.00$\pm$0.00                          & 21.2$\pm$0.82           & \textbf{4.96$\pm$0.81} \\
                     & VO   & 1                           & \cellcolor[HTML]{FFCCC9}0.80$\pm$1.40  & 27.28$\pm$7.78          & 12.70$\pm$4.73         \\
                     & RVO  & 1                           & 0.00$\pm$0.00                          & \textbf{19.76$\pm$1.40} & 18.11$\pm$0.70         \\
                     & hVO  & \cellcolor[HTML]{FFCCC9}0   & 0.00$\pm$0.00                          & 0.73$\pm$0.22           & 6.99$\pm$0.26          \\
\multirow{-5}{*}{8}  & OVVO & \cellcolor[HTML]{FFCCC9}0.6 & \cellcolor[HTML]{FFCCC9}3.50$\pm$4.50  & 29.69$\pm$16.43         & 48.09$\pm$3.13         \\ \midrule
                     & Ours & 1                           & 0.00$\pm$0.00                          & 25.38$\pm$1.91          & \textbf{6.18$\pm$0.06} \\
                     & VO   & 1                           & \cellcolor[HTML]{FFCCC9}15.00$\pm$6.78 & 33.24$\pm$5.37          & 17.30$\pm$6.37         \\
                     & RVO  & 1                           & \cellcolor[HTML]{FFCCC9}0.30$\pm$0.95  & \textbf{23.43$\pm$1.98} & 24.94$\pm$1.12         \\
                     & hVO  & \cellcolor[HTML]{FFCCC9}0   & 0.00$\pm$0.00                          & 0.65$\pm$0.11           & 7.90$\pm$0.60          \\
\multirow{-5}{*}{12} & OVVO & \cellcolor[HTML]{FFCCC9}0   & \cellcolor[HTML]{FFCCC9}13.80$\pm$5.63 & 29.80$\pm$26.06         & 70.40$\pm$3.68         \\ \bottomrule
\end{tabular}
\caption{Results of the experiments. The entries marked in red have significant failures in terms of either success rate (S.R., reaching the goal) or collisions. Out of the non-red rows, the best in terms of simulation time and computation time are written in bold font. N.A -- Number of Agents. Simulation time -- average total episode length. Computation time -- per iteration solve and sampling time, respectively.}
\label{tab:results_experiments}
\end{table}



\section{Conclusion}
\label{sec_conclusions}
In this paper we have shown how to combine CBFs with VO in a way that guarantees safety, and leverages the efficiency provided by the VO by including it in the objective of the optimization, overcoming the downside of being overly conservative that results from
considering VOs as hard constraints.
The simulation comparisons show that the proposed approach outperforms the alternatives in terms of path smoothness, success rates, and collision rates.







\balance
\bibliographystyle{IEEEtran}
\bibliography{main,MyLibraryPetter}

\end{document}